%% file: iclr2027_conference.tex
\documentclass{article} % For LaTeX2e
\usepackage{iclr2027_conference,times}
\usepackage{graphicx}

\newcommand{\Keep}{\textsc{Keepsake}\xspace}

\usepackage{amsmath,amssymb}
\usepackage{algorithm}
\usepackage{algpseudocode}
\usepackage{booktabs}
\usepackage{wrapfig}
\usepackage{tabularx,array}
\usepackage{needspace}
\usepackage{subcaption}
\usepackage{pgfplots}
\usepgfplotslibrary{groupplots}
\pgfplotsset{compat=1.18}
\graphicspath{{figures/}{./}}
\definecolor{memcamblue}{RGB}{52,114,245}
\definecolor{fifoorange}{RGB}{230,138,0}
\definecolor{keepgreen}{RGB}{34,153,84}
\definecolor{lightbox}{RGB}{245,247,250}
\definecolor{boxborder}{RGB}{180,185,195}
\input{math_commands.tex}

\usepackage{hyperref}
\usepackage{url}
\usepackage{xspace}

\title{\textbf{Keepsake}: Selective Spatial Memory for Long-Horizon Video Generation}

\iclrfinalcopy 
\author{%
Abdul Mohaimen Al Radi \quad Kunyang Li \quad Yuzhang Shang \quad Mubarak Shah \quad Yu Tian\\
Institute of Artificial Intelligence, University of Central Florida\\
\texttt{\{ab575577, kunyang.li, yuzhang.shang, yu.tian2\}@ucf.edu, shah@crcv.ucf.edu}
}

\begin{document}

\maketitle
\begin{abstract}
Long-horizon camera-controlled video generation relies on persistent memory to maintain scene consistency beyond the model's immediate temporal context. Existing systems follow two strategies to achieve this consistency. Full-history approaches retain all generated observations, causing unbounded storage and retrieval costs to grow while allowing low-fidelity memories to persist. Selective-construction approaches reduce redundancy, but make one-time retention decisions that are never revisited, even as an observation’s value changes with the evolving memory bank. Both strategies leave a shared question unresolved: as the generated history evolves, which stored observations should still remain in memory? Our key insight is that the value of a stored observation is not fixed, but relational: it depends on the alternatives currently available in the memory bank. A view supported by many geometrically and visually similar substitutes can be relinquished with little loss of coverage, whereas an observation with few viable alternatives should remain regardless of age. We introduce \Keep{}, an online, training-free controller for fixed-capacity spatial memory. At each update, \Keep{} constructs a pose–appearance graph over retained and newly generated observations, combining camera-pose proximity with visual similarity. A retention priority jointly captures the number of strong substitutes and the similarity of the closest alternative, allowing \Keep{} to continually reassess memory value, preserve observations with little alternative support, and evict highly replaceable ones under a fixed budget. The controller modifies only the persistent-memory update; the host generator, denoising schedule, and retrieval rule remain unchanged. Across MemCam and WorldMem, \Keep{} improves FVD and LPIPS under a fixed memory budget; on 180-second MemCam trajectories, it retains only 32 of 5,397 frames while reducing FVD by 35.1\%.
\begin{figure}[!ht]
    \centering
    \includegraphics[width=\linewidth]{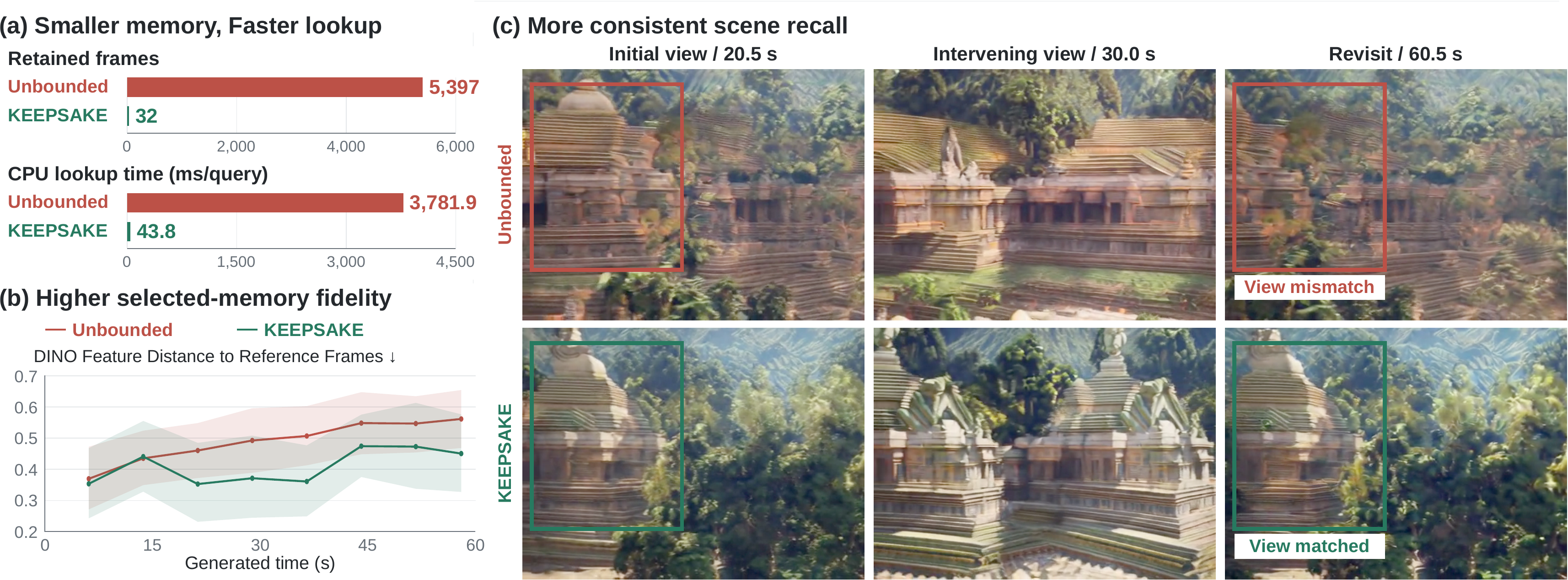}
    %\vspace{-1.5em}
    \caption{
    \textbf{What should remain in memory?}
    Unbounded memory versus our budgeted memory in MemCam.
    (a) Retained-frame count and CPU lookup latency per query.
    (b) DINO feature cosine distance from selected historical frames to
    reference frames over generation time; lower is better.
    (c) A camera-revisit example, with boxes highlighting corresponding
    scene regions.}
    \label{fig:motivation}
    %\vspace{-2em}
\end{figure}
\end{abstract}

\section{Introduction}
Camera-controlled video generation enables deliberate viewpoint navigation within generated scenes, supporting interactive exploration and controllable content creation. Extending this capability to long horizons requires preserving scene consistency beyond the model's immediate temporal context, particularly when previously observed regions are revisited after substantial temporal gaps. 
Recent memory-augmented methods address this challenge by maintaining historical observations associated with camera states, retrieving relevant evidence for subsequent generation, and writing newly generated observations back into persistent memory \cite{memcam,worldmem,contextasmemory,vmem}.  Each generation step is therefore both an output and a potential source of evidence for future generation.
%Each generated observation can therefore serve as evidence for future generation.

%Long-horizon video generation aims to maintain coherent visual content over sequences that substantially exceed a model's immediate temporal context. This challenge is particularly pronounced in camera-controlled generation, where previously observed regions may become relevant again after long temporal gaps. Memory-augmented generators address this by operating in a closed loop: before synthesizing each chunk, a retriever selects historical observations relevant to the upcoming trajectory; the generator conditions on this context; and newly generated frames are written back into a persistent memory associated with camera poses \cite{memcam, worldmem, contextasmemory, vmem}. Each generation step is therefore both an output and a potential source of evidence for future generation.

Existing spatial-memory systems commonly retain the full generation history and rely on retrieval to select relevant observations at query time \cite{memcam,worldmem}. This strategy introduces two coupled limitations. First, the memory archive grows continuously with generation, leading to increasing retrieval overhead as the number of stored observations expands (Fig.~\ref{fig:motivation}(a)). Second, quality degradation can persist in the memory archive, allowing drifted observations to be repeatedly retrieved and reused. As shown in Fig.~\ref{fig:motivation}(b), memories with well-matched viewpoints can still deviate substantially from the established scene, indicating that geometric relevance alone does not guarantee reliable memory. Fig.~\ref{fig:motivation}(c) further shows the resulting inconsistency when the camera revisits the same region, where the regenerated scene no longer matches its earlier appearance.

%Existing spatial-memory systems follow two strategies. Full-history approaches retain all generated observations and rely on retrieval to select useful context at query time \cite{memcam, worldmem}. This creates two coupled problems. First, the archive grows with generated history, so the candidate set that exhaustive retrieval must score at each query expands continuously, even when the number of memories ultimately supplied to the generator remains fixed \ref{tab:exp-quality}. Second, the closed-loop nature of generative memory means that retrieval errors can propagate: retrieved memories condition generation, those outputs are written back into the archive, and degraded evidence therefore remains eligible to influence future chunks. Figure \ref{fig:motivation}(b) shows this happening in practice: as generation continues, retrieval keeps finding views that match the camera angle well, but those views are increasingly the model's own earlier, drifted outputs rather than anything close to the real scene. A well-matched memory is not a useful one if its content has already degraded.

Selective-construction approaches reduce redundant accumulation through spatial admission, keyframe selection, or memory merging \cite{vmem,anchor,mosaic}. However, these decisions are typically made only when an observation is written and are not revisited as the memory evolves. This is limiting because the value of a stored observation is inherently dynamic: a view that is unique at admission may later become redundant as similar evidence accumulates, while an older view with few substitutes may remain important. This raises a shared question across existing memory strategies: as generation proceeds, which observations should still remain in memory?

We formulate this as a dynamic reassessment problem. The retention value of an observation is relational rather than intrinsic, depending on the alternatives currently available in memory. A view with several geometrically and visually similar substitutes can be removed with limited loss of coverage, whereas an observation with few viable alternatives should be retained regardless of its age or admission-time novelty. Selective forgetting therefore serves not only to bound memory growth, but also to continually reallocate capacity toward observations that remain most informative.

We introduce \Keep{}, an online, training-free, plug-and-play memory controller that dynamically reassesses which observations should remain in memory. At each update, \Keep{} pools the retained memory with newly generated observations and constructs a pose–appearance graph whose edge weights combine camera-pose proximity with visual similarity. Each observation is scored by its number of strong substitutes and the similarity of its closest alternative. Low-priority observations, those with many strong substitutes are evicted until a fixed budget $B$ is satisfied. Crucially, retention priorities depend on the current candidate set rather than being fixed at admission, so an observation that was previously important can become replaceable as new alternatives arrive. 
%The controller modifies only the persistent-memory update. The underlying generator, denoising schedule, and retrieval rule remain unchanged, and no retraining or scene reconstruction is required. For fixed budget and chunk size, archive size and per-query retrieval candidates are bounded independently of video duration.
The fixed budget bounds memory growth and retrieval overhead, while dynamic reassessment continually reallocates capacity toward observations that remain most useful.

Our experiments show selective forgetting can improve generation quality while substantially reducing memory size. On fifteen matched 180-second MemCam \cite{memcam} trajectories, \Keep{} retains only 32 observations instead of 5,397 while reducing FVD from 734.2 to 476.6, a 35.1\% improvement. On fifteen matched 60-second WorldMem \cite{worldmem} videos, FVD decreases from 3077.6 to 1116.9. Importantly, simply imposing a memory cap is not sufficient: FIFO retention degrades generation quality relative to full retention, and different bounded policies yield substantially different results under the same budget (Tab.~\ref{tab:exp-quality}). These results show memory quality depends not only on how much history is retained, but also on which observations remain available.

Our contributions can be summarized as follows:
\begin{itemize}
    \item We identify a fundamental limitation of growing spatial memory in camera-controlled long-horizon video generation and formulate retention as a dynamic reassessment problem: an observation's value depends on the substitutes currently available in memory, rather than its age or admission-time novelty.
    \item We propose \Keep{}, a training-free memory controller that continuously reassesses retention value through pose--appearance substitutability, preserving observations with few viable alternatives under a fixed budget.
    \item Experiments on MemCam and WorldMem show that selective forgetting can simultaneously improve long-horizon generation quality and substantially reduce memory and retrieval cost, outperforming both full-history retention and bounded baselines.
\end{itemize}

\section{Related Work}
\label{sec:related-work}
%\syz{we should include more reference. now we only include 9 papers at related work.}
\paragraph{Camera-Controlled Video Generation.}
Recent text-to-video and image-to-video models have achieved impressive visual quality and realism \cite{chen2024videocrafter2,yang2025cogvideox,kong2024hunyuanvideo,blattmann2023stable}. However, they offer limited control over viewpoint evolution. Explicit camera control is therefore essential for applications that require deliberate viewpoint navigation and interactive exploration of a generated scene. Accordingly, recent studies directly condition video generation on camera trajectories \cite{wang2024motionctrl, yang2024direct,he2024cameractrl,zheng2024cami2v,hou2024training}.
\cite{yang2024direct, wang2024motionctrl} disentangle camera motion from object motion, while \cite{he2024cameractrl} introduce explicit camera-pose conditioning into pretrained video generators. More recent work further improves camera controllability through geometric and correspondence priors, including epipolar constraints, trajectory-based correspondence, monocular depth, and explicit 3D scene representations \cite{zheng2024cami2v, hou2024training,xiao2025trajectory,li2502realcam, ren2025gen3c}. %Collectively, these advances improve the precision and geometric consistency of camera-controlled video generation. 
Building on camera-controllable video generation,  \Keep{} focuses on maintaining scene consistency over long horizons, particularly preserving previously observed content when the camera revisits earlier viewpoints after extended exploration.

\paragraph{Long-Horizon Consistency under Camera Control.} 
Long-horizon camera exploration requires preserving scene information beyond the model's local context, particularly when previously observed regions are revisited after large temporal or viewpoint gaps. Recent methods address this challenge with persistent spatial memory, storing historical observations and retrieving evidence relevant to the current viewpoint \cite{worldmem,contextasmemory,memcam}. Other approaches organize memory through geometry-aware indexing, spatial correspondence, local spatial memories, or persistent 3D representations to better preserve cross-view consistency \cite{vmem,ren2025gen3c,xu2026ucm,zhang2026worldstereo,zhaospatia,anchor,mosaic,hong2025relic}. Beyond full-history memory, long-video generation has also explored adaptive memory selection and bounded retention under limited context budgets \cite{memflow,surprise,li2026packcache}. However, these methods either retain an expanding archive or make retention decisions primarily at admission time. \Keep{} further studies how memory should be maintained as it grows, showing that retaining more historical observations is not always beneficial. It selectively preserves observations that remain useful for future revisits, improving revisit consistency while controlling storage and computation.

\section{Keepsake: Memory Budgeting Method}
\label{sec:method}

\begin{figure}[t]
    \centering
    \includegraphics[width=\linewidth]{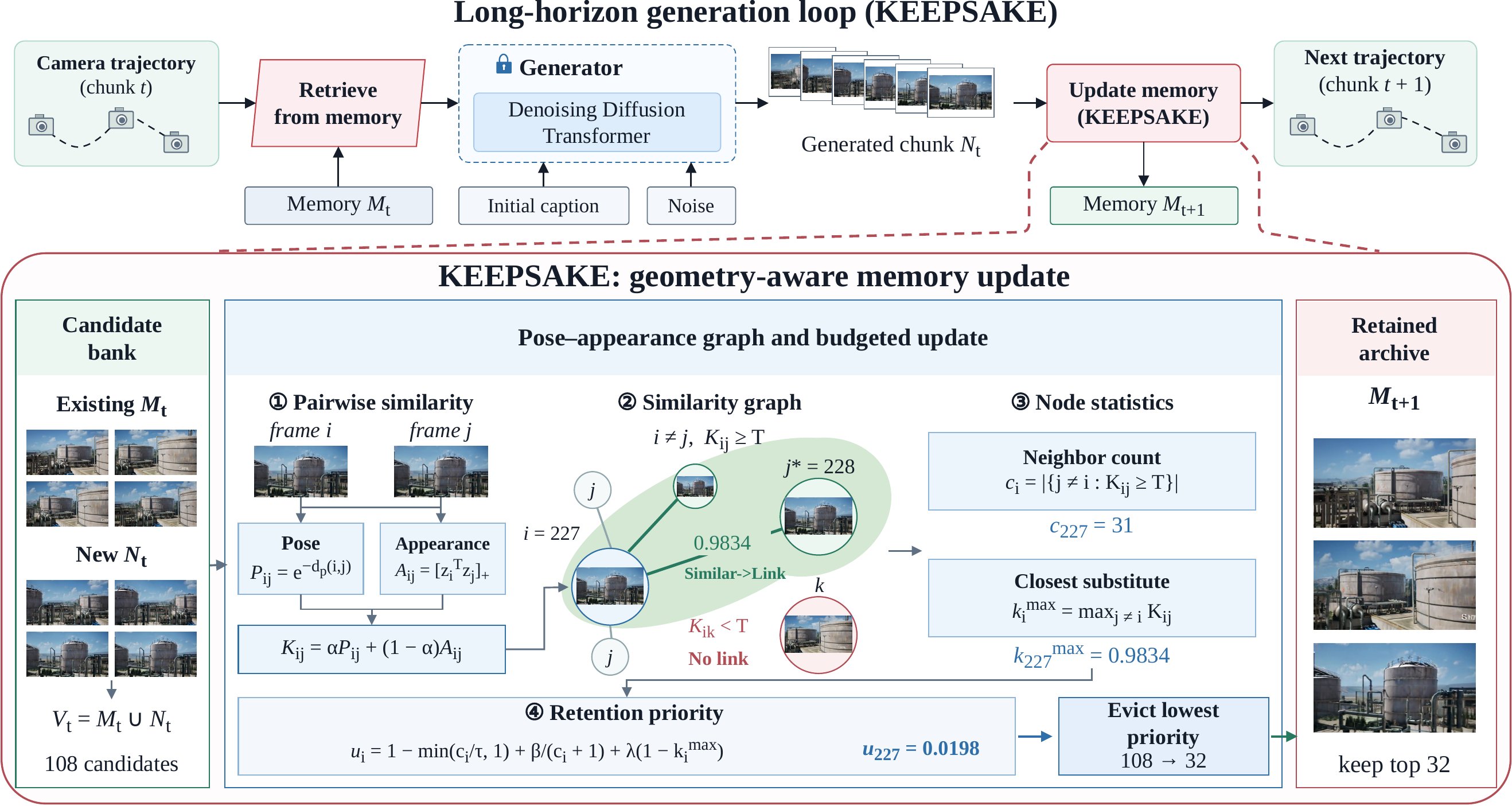}
    \caption{
    \textbf{Overview of \Keep{}.}
    \textbf{Top:} \Keep{} is inserted into the long-horizon generation loop as a memory-update module. After each generated chunk, it updates the persistent memory while leaving the generator and retrieval rule unchanged.
    \textbf{Bottom:} Existing memory and newly generated observations are pooled into a candidate bank and organized as a pose-appearance graph, where edge weights combine camera-pose proximity and visual similarity. Node statistics capture both the number of strong substitutes and the similarity to the closest substitute, which jointly determine retention priority. Low-priority observations are evicted until the fixed budget $B$ is met. Thumbnails and displayed scores are illustrative.}

    \label{fig:keepsake-method}
\end{figure}

\Keep{} maintains a memory bank of at most $B$ observations by
reassessing redundancy as new views arrive. The retention priority
of an image depends on the alternatives currently available:
how many views provide similar evidence, and how closely its strongest
alternative matches it. We capture these relationships through a
pose-appearance graph that combines camera geometry with
visual similarity. At each update, the controller ranks retained and
incoming images, favoring views with limited alternative support
and removing low-priority observations to satisfy the budget. We pose memory retention as a prospective policy, as in interactive video world models, there is no certainty on where the camera will move next. Figure~\ref{fig:keepsake-method} illustrates this update within the
existing generation loop; the generator and retrieval rule remain
unchanged. In this section, we cover how our strict memory budgeting is enforced and updated at each step along with construction of the pose-appearance graph.

\subsection{Fixed-Budget Spatial Memory}
\label{sec:memory-interface}

At step $t$, the generator produces a video chunk $\hat{Y}_t$ along
a prescribed camera trajectory $P_t$. Let $M_{t-1}$ denote the memory
bank available before generation, and let $B$ be its capacity in
stored observations. The retriever $R$ selects historical context
$C_t$ according to the upcoming camera poses. The generator $G$
then synthesizes the next chunk using this context and the preceding
chunk $\hat{Y}_{t-1}$:
\begin{equation}
    C_t = R(P_t;M_{t-1}),
    \qquad
    \hat{Y}_t = G(\hat{Y}_{t-1},P_t,C_t).
    \label{eq:memory-generation-loop}
\end{equation}

After generation, the host constructs $N_t$, the set of new memory
observations obtained from $\hat{Y}_t$. In the image-memory setting,
each observation contains a generated frame, its camera pose, and
an appearance descriptor. These observations are combined with
the existing memory bank to form the candidate set
$V_t=M_{t-1}\cup N_t$. \Keep{} determines which candidates remain:
\begin{equation}
    M_t = U(M_{t-1},N_t;B),
    \qquad
    M_t\subseteq V_t,
    \qquad
    |M_t|\leq B.
    \label{eq:memory-update}
\end{equation}
Only the update rule $U$ changes; the generator and retriever remain
unchanged. The budget limits the persistent memory bank, rather
than merely the context retrieved for a single generation step.

\subsection{Pose-Appearance Graph}
\label{sec:pose-appearance}

We construct a graph whose nodes are the candidate observations in
$V_t$. Each observation $o_i$ is associated with a camera-to-world
pose $(R_i,\mathbf{t}_i)$ and a normalized visual descriptor
$\mathbf{z}_i$. Here, $R_i$ and
$\mathbf{t}_i$ denote camera rotation and position, respectively.
The edge weight $K(i,j)$ combines camera-pose proximity with
descriptor similarity: the former measures closeness in viewing
position and direction, while the latter captures visual resemblance.
Together, these signals provide a proxy for shared view coverage,
which we use to assess redundancy within the memory bank.

\paragraph{Pose distance.}
We measure camera-pose separation by combining translation and rotation:
\begin{equation}
    d_p(i,j)
    =
    \frac{\|\mathbf{t}_i-\mathbf{t}_j\|_2}
         {\max(\widetilde{d}_t,\epsilon)}
    +
    \frac{2\theta_{ij}}{\pi},
    \label{eq:pose-distance}
\end{equation}
where $\widetilde{d}_t$ is the median nonzero pairwise translation
distance among observations in $V_t$, and $\theta_{ij}\in[0,\pi]$
is the rotation angle of $R_i^\top R_j$. The translation normalization
expresses positional separation relative to the current candidate set,
while the rotation term accounts for differences in viewing direction.
We use $\widetilde{d}_t=1$ when all camera positions coincide and
$\epsilon=10^{-6}$ as a numerical floor.

\paragraph{Edge weights.}
We combine pose proximity and appearance similarity to define the
weight between distinct observations:
\begin{equation}
    K(i,j)
    =
    \alpha\,e^{-d_p(i,j)}
    +
    (1-\alpha)\max(\mathbf{z}_i^\top\mathbf{z}_j,0),
    \qquad i\neq j,
    \label{eq:edge-weight}
\end{equation}
where $\alpha\in[0,1]$ controls the relative contribution of geometry
and appearance. The exponential converts pose distance into similarity.
Since the visual descriptors are normalized, their inner product
measures cosine similarity, with negative values clipped to zero.
The resulting weights are symmetric and lie in $[0,1]$. Larger
weights indicate greater similarity between observations, providing
the basis for measuring redundancy within the memory bank.

\paragraph{Node degree and maximum similarity.}
To assess redundancy, we compute two statistics for each observation:
its degree in the thresholded graph, which counts strongly similar
observations, and its maximum similarity to any other candidate.
Thresholding the edge weights at $T\in[0,1]$ defines an undirected
graph $\mathcal{G}_t=(V_t,E_t)$, with
\begin{equation}
    E_t =
    \left\{
        \{o_i,o_j\} : i<j,\; K(i,j)\geq T
    \right\}.
    \label{eq:graph-edges}
\end{equation}
For each observation, we compute its degree in this graph and
the weight of its strongest alternative:
\begin{equation}
    c_i =
    \sum_{j\neq i}\mathbb{I}[K(i,j)\geq T],
    \qquad
    k_i^{\max} = \max_{j\neq i}K(i,j).
    \label{eq:substitute-statistics}
\end{equation}
The degree $c_i$ counts observations with high geometric--visual
similarity, while $k_i^{\max}$ measures the similarity of the
closest alternative. These statistics distinguish how many
strong alternatives are available from how closely the best
one matches the observation. The maximum is computed over all
other candidates in $V_t$, including those whose weights fall
below $T$.
\subsection{Retention Priority}
\label{sec:retention-utility}

We combine node degree and maximum similarity into a retention
priority. Low-degree observations have fewer strongly similar
alternatives, while a large maximum similarity indicates a close
match elsewhere in the candidate set. We therefore assign:
\begin{equation}
    \begin{aligned}
        u_i &= \phi(c_i) + \lambda(1-k_i^{\max}), \\
        \phi(c) &= 1-\min\!\left(\frac{c}{\tau},1\right)
                  +\frac{\beta}{c+1},
    \end{aligned}
    \label{eq:retention-utility}
\end{equation}
where higher $u_i$ indicates greater retention priority.
The degree-based function $\phi$ decreases as an observation
acquires more strongly similar alternatives. Its linear term
reaches zero at degree $\tau$, while the inverse-degree term
continues to distinguish nodes beyond this threshold.
The remaining term, $1-k_i^{\max}$, increases priority when
even the closest alternative is dissimilar. The parameters
$\beta\geq0$ and $\lambda\geq0$ control the contributions of
inverse degree and maximum similarity, respectively, with
$\tau>0$ specifying the degree threshold.

Retention priorities depend on the current candidate set $V_t$,
rather than being fixed when an observation first enters memory.
New observations can provide additional alternatives for an existing
view, reducing its priority, while older views with few alternatives
remain favored. When the candidate set exceeds the budget, \Keep{}
computes all priorities once and uses this fixed ranking throughout
eviction. Node degrees and maximum similarities are not recomputed
after individual removals.

\subsection{Budgeted Memory Update}
\label{sec:budgeted-eviction}

At each update, \Keep{} evaluates retention priorities over the
candidate set $V_t=M_{t-1}\cup N_t$. When $|V_t|>B$, it removes
the $|V_t|-B$ lowest-priority observations eligible for eviction,
breaking equal-priority ties by earlier insertion. Otherwise,
all observations are retained. Priorities are computed once per
update and remain fixed during removal in Algorithm \ref{alg:keepsake}. The resulting memory
bank supplies candidates to the original retriever, without
changing its ranking rule.

\begin{algorithm}[]
    \caption{\Keep{}: budgeted memory update}
    \label{alg:keepsake}
    \begin{algorithmic}[1]
        \Require Memory bank $M_{t-1}$, new observations $N_t$,
                 budget $B$
        \Ensure Updated memory bank $M_t$, with $|M_t|\leq B$

        \State $V_t \gets M_{t-1}\cup N_t$
        \If{$|V_t|\leq B$}
            \State \Return $V_t$
        \EndIf
        \State Compute edge weights $K(i,j)$ over $V_t$
               using Eq.~(\ref{eq:edge-weight})
        \State Compute degrees $c_i$ and maximum similarities
               $k_i^{\max}$ using
               Eq.~(\ref{eq:substitute-statistics})
        \State Compute retention priorities $u_i$ once
               using Eq.~(\ref{eq:retention-utility})
        \State Rank eviction-eligible observations by increasing
               $u_i$, breaking ties by earlier insertion
        \State $M_t \gets V_t$
        \State Remove the first $|V_t|-B$ ranked observations
               from $M_t$
        \State \Return $M_t$
    \end{algorithmic}
\end{algorithm}

\subsection{Computational Cost}
\label{sec:resource-bounds}
Let $s$ denote the maximum number of new observations introduced per chunk. Since \Keep{} retains at most $B$ observations, each update considers at most $B+s$ candidates. Computing pairwise pose--appearance similarities and retention priorities requires $O((B+s)^2)$ time, while sorting adds $O((B+s)\log(B+s))$. The retained archive uses $O(B)$ storage, and exhaustive retrieval scores at most $B$ candidates per query. Therefore, for fixed $B$ and chunk size, memory usage, update cost, and retrieval cost remain bounded independently of video duration. \Keep{} modifies only the persistent-memory update and requires no generator retraining or changes to the retrieval rule.

\providecommand{\Keep}{\textsc{Keepsake}}

% \section{Experiments}
\section{Experiments}
\label{sec:experimental-results}

We investigate how much spatial memory long-horizon generation needs
and how that capacity should be allocated. Our primary comparisons
use MemCam and WorldMem, varying memory retention while keeping
the generator and retrieval rule unchanged within each system.
Comparisons against complete retention test whether a compact memory
bank can preserve generation quality; comparisons at matched budgets
test whether the choice of retained observations matters.

\subsection{Experimental Setup}
\label{sec:exp-protocol}

\paragraph{Generation systems and datasets.}
We evaluate \Keep{} in two settings: MemCam~\citep{memcam} on the
Context-as-Memory dataset~\citep{contextasmemory}, and
WorldMem~\citep{worldmem} on its Minecraft world-simulation dataset.
Each primary evaluation contains fifteen matched 60-second videos.
A separate set of fifteen matched 180-second MemCam videos also
test longer generation. Within each system, retention policies
share the initial inputs, prescribed control sequences, generator
checkpoint, generation settings, and random seeds. We report
comparisons separately for each system and duration, so that
differences between retention policies are measured under the
same generation conditions.

\paragraph{Retention baselines.}
We compare \Keep{} with complete retention and three bounded baselines: FIFO, K-center, and Marginal Coverage Eviction (MCE). FIFO prioritizes recency by discarding the oldest eligible observations. K-center selects representatives to improve worst-case coverage, while MCE removes observations whose deletion minimally reduces coverage over representative historical queries. Both coverage-based baselines use geometry and appearance, whereas \Keep{} ranks observations by redundancy within the current candidate bank using node degree and maximum similarity. All bounded methods use the same archive budget, with $B=32$ in the main comparison. Additional algorithmic details are provided in the appendix~\ref{app:ks-baseline-algorithms}.

\paragraph{Implementation details.}
We use MemCam with the Wan2.1 1.3B \cite{wan} backbone, generating 76-frame sections with 50 denoising steps. For retention scoring, generated
images are represented by $\ell_2$-normalized DINOv2-Base \cite{dinov2} descriptors.

We use WorldMem's released Minecraft diffusion and VAE checkpoints,
generating frames at $128\times128$ resolution with 20 DDIM steps
and $\eta=0$. For retention scoring, predicted VAE latents are
spatially average-pooled to $4\times4$, flattened, and $\ell_2$-normalized.

Both systems share the default retention parameters:
$\alpha=0.65$ for pose-appearance weighting, $T=0.65$ for edge
construction, and $(\tau,\beta,\lambda)=(3,0.5,0.25)$ for
retention priority. We evaluate budgets
$B\in\{16,32,64,128\}$, using $B=32$ for the main comparison.

\paragraph{Evaluation metrics.}
We measure frame-level fidelity using LPIPS~\citep{lpips}, comparing
generated frames with ground truth at the corresponding trajectory
indices, and video-level distributional quality using FVD~\citep{fvd}.
Lower values indicate better performance for both metrics.
We additionally evaluate six VBench~\citep{vbench} dimensions:
subject consistency, background consistency, motion smoothness,
dynamic degree, aesthetic quality, and imaging quality.
These dimensions characterize perceptual quality and temporal
behavior; we report them individually on a $\times100$ scale
rather than constructing an aggregate score.

\subsection{Generation Quality under Bounded Memory}
\label{sec:exp-quality}

Table~\ref{tab:exp-quality} compares complete retention with
bounded-memory policies on the 60-second evaluation suites.
With $B=32$, \Keep{} improves both LPIPS and FVD over complete
retention in both systems. In MemCam, LPIPS decreases from
0.5916 to 0.5850 and FVD from 784.9 to 690.5. In WorldMem,
the corresponding improvements are 0.6523 to 0.5337 and
3077.6 to 1116.9. Among budget-matched policies, \Keep{}
achieves the lowest LPIPS in both systems and the lowest
FVD in WorldMem. FIFO, in contrast, worsens both metrics relative
to complete retention in both systems. These results show
that compact memory can improve generation quality, but
imposing a capacity limit alone is insufficient: different
retention rules yield substantially different outcomes
under the same budget.
% table 1
\begin{table}[!t]
\centering
\caption{
Main quality results on 15 matched 60-second videos within each system.
All bounded methods use $B=32$. For consistency, we report FVD and LPIPS. For efficiency, we report number of stored items in memory and retrieval time needed for the model for each generation chunk. For additional evaluation we also report Vbench scores across subject and background consistency, as well as aesthetics and average scores.
}
\label{tab:exp-quality}
\setlength{\tabcolsep}{4pt}
\renewcommand{\arraystretch}{1.15}
\resizebox{\textwidth}{!}{%
\begin{tabular}{lrrrrrrrr}
\toprule
& \multicolumn{2}{c}{Consistency}
& \multicolumn{4}{c}{VBench $\uparrow$}
& \multicolumn{2}{c}{Efficiency} \\
\cmidrule(lr){2-3}
\cmidrule(lr){4-7}
\cmidrule(lr){8-9}
Models
& FVD $\downarrow$
& LPIPS $\downarrow$
& Subject
& Background
& Aesthetic
& Average
& \shortstack{Stored\\items $\downarrow$}
& \shortstack{Retrieval\\ms/query $\downarrow$} \\
\midrule
MemCam
& 784.9 & 0.59 & 80.53 & 90.16 & 45.42 & 73.98
& 1825 & 1344.61 \\
MemCam + FIFO
& 832.2 & 0.62 & 77.01 & 89.50 & 45.35 & 73.34
& 32 & \textbf{39.93} \\
MemCam + MCE
& 773.9 & 0.60 & 79.78 & 89.97 & 46.05 & 73.96
& 32 & 54.20 \\
MemCam + K-center
& 712.0 & 0.59 & 80.85 & 90.00 & 45.52 & 74.11
& 32 & 43.59 \\

MemCam + \textbf{KEEPSAKE (Ours)}
& \textbf{690.5} & \textbf{0.58} & \textbf{81.78} & \textbf{90.47}
& \textbf{46.14} & \textbf{74.65} & 32 & 43.77 \\
\midrule
WorldMem
& 3077.6 & 0.65 & 73.87 & 87.53 & 38.37 & 68.66
& 1,200 & 77.92 \\
WorldMem + FIFO
& 3554.9 & 0.69 & 72.47 & 86.15 & 36.92 & 68.61
& 32 & 4.670 \\
WorldMem + MCE
& 1924.0 & 0.57 & 79.12 & 90.30 & 41.54 & 72.67
& 32 & 4.71 \\
WorldMem + K-center
& 1419.6 & 0.56 & 79.60 & 91.58 & 43.04 & 72.84
& 32 & 5.00 \\

WorldMem + \textbf{KEEPSAKE (Ours)}
& \textbf{1116.9} & \textbf{0.53} & \textbf{81.71}
& \textbf{92.95} & \textbf{43.72} & \textbf{74.05}
& 32 & \textbf{4.54} \\
\bottomrule
\end{tabular}%
}
\end{table}

\paragraph{MemCam benchmark.}
We also evaluate \Keep{} using MemCam's original camera round-trip
protocol on Context-as-Memory~\citep{contextasmemory}, covering
$90^\circ$ (153-frame) and $360^\circ$ (609-frame) trajectories. We include Diffusion Forcing(DFoT) model \cite{dfot} and Geometry Forcing \cite{geometry} in the comparison as well.
As shown in Table~\ref{tab:native-benchmarks}, \Keep{} remains
competitive with a memory budget of $B=32$, achieving higher SSIM
and lower LPIPS than MemCam in both settings. On the $360^\circ$
round trip, PSNR also increases from 14.81 to 16.22\,dB.

\begin{figure}[t]
\centering
\pgfplotsset{keepsake panel axes/.style={
  scale only axis, at={(24pt,10mm)}, anchor=south west,
  width=\dimexpr\linewidth-30pt\relax, height=32mm
}}
\captionsetup[subfigure]{font=small,skip=4pt,justification=centering,singlelinecheck=false}
\begin{subfigure}[t]{0.32\linewidth}
\centering\vspace{0pt}
% Generated by paper/export_memory_diagnostics_tikz.py.
% Requires \usepackage{pgfplots} and \pgfplotsset{compat=1.18}.
% Input inside a subfigure; do not resizebox the result (fonts would shrink).
\begingroup
\definecolor{diagKeep}{HTML}{4C8C4A}
\definecolor{diagFifo}{HTML}{D55E00}
\definecolor{diagUnbounded}{HTML}{4D4D4D}
\begin{tikzpicture}
\path[use as bounding box] (0,0) rectangle (\linewidth,52mm);
\begin{axis}[
  keepsake panel axes,
  xmin=0.58, xmax=0.665, ymin=430, ymax=850,
  xtick={0.58,0.62,0.66}, ytick={450,600,750}, scaled ticks=false,
  xlabel={LPIPS $\downarrow$},
  tick label style={font=\scriptsize}, label style={font=\small},
  axis lines=left, axis line style={black!60}, tick style={black!60},
  tick align=outside, grid=major, grid style={black!12,line width=0.25pt},
  clip=false, enlargelimits=false,
]
\node[anchor=south west,font=\small,inner sep=0pt]
  at (rel axis cs:0,1.04) {FVD $\downarrow$};
\addplot[only marks,mark=*,mark size=2.6pt,diagUnbounded]
  coordinates {(0.5980,734.2)};
\node[anchor=south west,align=left,font=\scriptsize,text=diagUnbounded,
      inner sep=0pt,xshift=3pt,yshift=3pt]
  at (axis cs:0.5980,734.2) {MemCam\\[-1pt](0.5980, 734.2)};
\addplot[only marks,mark=square*,mark size=2.4pt,diagFifo]
  coordinates {(0.6514,677.3)};
\node[anchor=north east,align=right,font=\scriptsize,text=diagFifo,
      inner sep=0pt,xshift=-3pt,yshift=-4pt]
  at (axis cs:0.6514,677.3) {FIFO\\[-1pt](0.6514, 677.3)};
\addplot[only marks,mark=diamond*,mark size=2.8pt,diagKeep]
  coordinates {(0.5876,476.6)};
\node[anchor=south west,align=left,font=\scriptsize,text=diagKeep,
      inner sep=0pt,xshift=4pt,yshift=4pt]
  at (axis cs:0.5876,476.6) {KEEPSAKE (Ours)\\[-1pt](0.5876, 476.6)};
\end{axis}
\end{tikzpicture}
\endgroup
\caption{Long-horizon generation quality.}
\label{fig:long-horizon-extension}
\end{subfigure}\hfill%
\begin{subfigure}[t]{0.32\linewidth}
\centering\vspace{0pt}
% Generated by paper/export_memory_diagnostics_tikz.py.
% Requires \usepackage{pgfplots} and \pgfplotsset{compat=1.18}.
% Input inside a subfigure; do not resizebox the result (fonts would shrink).
\begingroup
\definecolor{diagKeep}{HTML}{4C8C4A}
\definecolor{diagFifo}{HTML}{D55E00}
\definecolor{diagUnbounded}{HTML}{4D4D4D}
\begin{tikzpicture}
\path[use as bounding box] (0,0) rectangle (\linewidth,52mm);
% Matched 60-s / 15-trajectory comparison; common baseline pixels.
\begin{axis}[
  keepsake panel axes,
  xmin=0.000000, xmax=0.618547, ymin=-0.28, ymax=1.56,
  ytick=\empty, xtick distance=0.2, scaled ticks=false,
  tick label style={font=\scriptsize}, label style={font=\small},
  xlabel={Feature distance $\downarrow$}, axis lines=left,
  y axis line style={draw=none}, axis line style={black!60}, tick style={black!60},
  tick align=outside, xmajorgrids=true, grid style={black!12,line width=0.25pt},
  clip=false, enlargelimits=false,
  legend style={font=\scriptsize,draw=none,fill=none,at={(0,1.01)},anchor=south west,
                legend columns=1,inner sep=0pt,row sep=-1pt},
  legend cell align=left,
]
\addlegendimage{only marks,mark=*,diagUnbounded}
\addlegendentry{Unbounded}
\addlegendimage{only marks,mark=triangle*,diagFifo}
\addlegendentry{FIFO}
\addlegendimage{only marks,mark=square*,diagKeep}
\addlegendentry{KEEPSAKE (Ours)}
\node[anchor=west,font=\scriptsize,inner sep=1pt,fill=white]
  at (axis cs:0.000000,1.40) {View mismatch};
\node[anchor=west,font=\scriptsize,inner sep=1pt,fill=white]
  at (axis cs:0.000000,0.40) {Memory corruption};
\draw[diagUnbounded,line width=1.1pt] (axis cs:0.104065,1.18) -- (axis cs:0.147862,1.18);
\addplot[only marks,diagUnbounded,mark=*,mark size=2.4pt]
  coordinates {(0.125602,1.18)};
\draw[diagUnbounded,line width=1.1pt] (axis cs:0.401161,0.18) -- (axis cs:0.571633,0.18);
\addplot[only marks,diagUnbounded,mark=*,mark size=2.4pt]
  coordinates {(0.487087,0.18)};
\draw[diagFifo,line width=1.1pt] (axis cs:0.133619,1.00) -- (axis cs:0.177677,1.00);
\addplot[only marks,diagFifo,mark=triangle*,mark size=2.4pt]
  coordinates {(0.155742,1.00)};
\draw[diagFifo,line width=1.1pt] (axis cs:0.419463,0.00) -- (axis cs:0.583547,0.00);
\addplot[only marks,diagFifo,mark=triangle*,mark size=2.4pt]
  coordinates {(0.502168,0.00)};
\draw[diagKeep,line width=1.1pt] (axis cs:0.257853,0.82) -- (axis cs:0.396434,0.82);
\addplot[only marks,diagKeep,mark=square*,mark size=2.4pt]
  coordinates {(0.325372,0.82)};
\draw[diagKeep,line width=1.1pt] (axis cs:0.315901,-0.18) -- (axis cs:0.498888,-0.18);
\addplot[only marks,diagKeep,mark=square*,mark size=2.4pt]
  coordinates {(0.407863,-0.18)};
\end{axis}
\end{tikzpicture}
\endgroup
\caption{View relevance and memory fidelity.}
\label{fig:retrieval-quality-paired}
\end{subfigure}\hfill%
\begin{subfigure}[t]{0.32\linewidth}
\centering\vspace{0pt}
% Generated by paper/export_memory_diagnostics_tikz.py.
% Requires \usepackage{pgfplots} and \pgfplotsset{compat=1.18}.
% Input inside a subfigure; do not resizebox the result (fonts would shrink).
\begingroup
\definecolor{diagKeep}{HTML}{4C8C4A}
\definecolor{diagFifo}{HTML}{D55E00}
\definecolor{diagUnbounded}{HTML}{4D4D4D}
\begin{tikzpicture}
\path[use as bounding box] (0,0) rectangle (\linewidth,52mm);
% Exact original 180-s / 13-trajectory rounded points, with RI omitted.
\begin{axis}[
  keepsake panel axes,
  xmin=-0.007, xmax=0.220, ymin=0.025, ymax=0.239,
  xtick={0,0.10,0.20}, ytick={0.05,0.10,0.15,0.20},
  scaled ticks=false,
  tick label style={font=\scriptsize,/pgf/number format/fixed,/pgf/number format/precision=2},
  label style={font=\small},
  xlabel={Retention gap $\downarrow$},
  axis lines=left, axis line style={black!60}, tick style={black!60},
  tick align=outside, grid=major, grid style={black!12,line width=0.25pt},
  clip=false, enlargelimits=false,
]
\node[anchor=south west,font=\small,inner sep=0pt]
  at (rel axis cs:0,1.04) {Selection gap $\downarrow$};
\addplot[diagFifo,line width=1pt,mark=none,->] coordinates {(0.1872,0.0448) (0.1668,0.0646) (0.1429,0.0867) (0.1203,0.1084)};
\addplot[only marks,diagFifo,mark=*,mark size=2.3pt,
  mark options={draw=white,line width=0.35pt,fill=diagFifo}] coordinates {(0.1872,0.0448)};
\node[anchor=east,font=\scriptsize\bfseries,text=diagFifo,
  inner sep=0.7pt,fill=white,xshift=-4pt]
  at (axis cs:0.1872,0.0448) {B16};
\addplot[only marks,diagFifo,mark=square*,mark size=2.3pt,
  mark options={draw=white,line width=0.35pt,fill=diagFifo}] coordinates {(0.1668,0.0646)};
\node[anchor=west,font=\scriptsize\bfseries,text=diagFifo,
  inner sep=0.7pt,fill=white,xshift=4pt]
  at (axis cs:0.1668,0.0646) {B32};
\addplot[only marks,diagFifo,mark=triangle*,mark size=2.3pt,
  mark options={draw=white,line width=0.35pt,fill=diagFifo}] coordinates {(0.1429,0.0867)};
\node[anchor=west,font=\scriptsize\bfseries,text=diagFifo,
  inner sep=0.7pt,fill=white,xshift=4pt]
  at (axis cs:0.1429,0.0867) {B64};
\addplot[only marks,diagFifo,mark=diamond*,mark size=2.3pt,
  mark options={draw=white,line width=0.35pt,fill=diagFifo}] coordinates {(0.1203,0.1084)};
\node[anchor=west,font=\scriptsize\bfseries,text=diagFifo,
  inner sep=0.7pt,fill=white,xshift=4pt]
  at (axis cs:0.1203,0.1084) {B128};
\addplot[diagKeep,line width=1pt,mark=none,->] coordinates {(0.0669,0.1037) (0.0426,0.1470) (0.0366,0.1596) (0.0298,0.1783)};
\addplot[only marks,diagKeep,mark=*,mark size=2.3pt,
  mark options={draw=white,line width=0.35pt,fill=diagKeep}] coordinates {(0.0669,0.1037)};
\node[anchor=west,font=\scriptsize\bfseries,text=diagKeep,
  inner sep=0.7pt,fill=white,xshift=4pt]
  at (axis cs:0.0669,0.1037) {B16};
\addplot[only marks,diagKeep,mark=square*,mark size=2.3pt,
  mark options={draw=white,line width=0.35pt,fill=diagKeep}] coordinates {(0.0426,0.1470)};
\node[anchor=west,font=\scriptsize\bfseries,text=diagKeep,
  inner sep=0.7pt,fill=white,xshift=4pt]
  at (axis cs:0.0426,0.1470) {B32};
\addplot[only marks,diagKeep,mark=triangle*,mark size=2.3pt,
  mark options={draw=white,line width=0.35pt,fill=diagKeep}] coordinates {(0.0366,0.1596)};
\node[anchor=east,font=\scriptsize\bfseries,text=diagKeep,
  inner sep=0.7pt,fill=white,xshift=-4pt]
  at (axis cs:0.0366,0.1596) {B64};
\addplot[only marks,diagKeep,mark=diamond*,mark size=2.3pt,
  mark options={draw=white,line width=0.35pt,fill=diagKeep}] coordinates {(0.0298,0.1783)};
\node[anchor=west,font=\scriptsize\bfseries,text=diagKeep,
  inner sep=0.7pt,fill=white,xshift=4pt]
  at (axis cs:0.0298,0.1783) {B128};
\addplot[only marks,diagUnbounded,mark=*,mark size=2.6pt]
  coordinates {(0,0.2188)};
\node[anchor=west,font=\scriptsize\bfseries,text=diagUnbounded,xshift=4pt,inner sep=0pt]
  at (axis cs:0,0.2188) {Unbounded};
\node[font=\scriptsize\bfseries,text=diagFifo,inner sep=1pt]
  at (axis cs:0.16,0.138) {FIFO};
\node[font=\scriptsize\bfseries,text=diagKeep,align=center,inner sep=1pt]
  at (axis cs:0.125,0.181) {KEEPSAKE\\(Ours)};
\end{axis}
\end{tikzpicture}
\endgroup
\caption{Retention-selection tradeoff.}
\label{fig:retention-selection}
\end{subfigure}
\caption{
    \textbf{Generation quality and memory diagnostics.}
    (a) LPIPS and FVD on fifteen matched 180-second MemCam rollouts.
    (b) View mismatch and memory corruption on fifteen matched 60-second rollouts, measured by DINOv2 cosine distance; whiskers denote 95\% trajectory-bootstrap confidence intervals. FIFO and \Keep{} use $B=32$ in (a,b).
    (c) Retention gap measures useful evidence lost through eviction, while selection gap measures the retriever's failure to select the best evidence that remains in memory. Results are shown on fifteen 180-second trajectories across memory budgets $B$. Lower is better for all metrics.
}
\label{fig:three-panel-placeholder}
\end{figure}
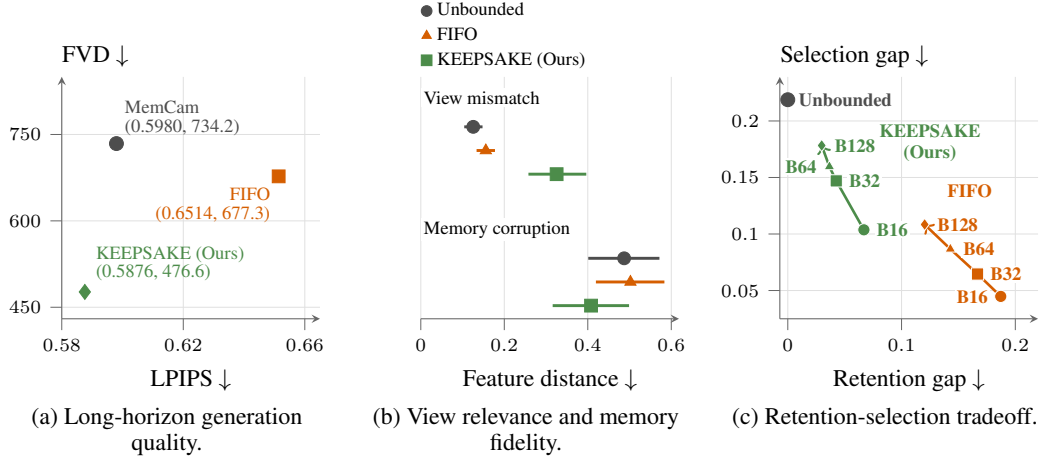

\begin{figure}[]
    \centering
    \includegraphics[width=\linewidth]{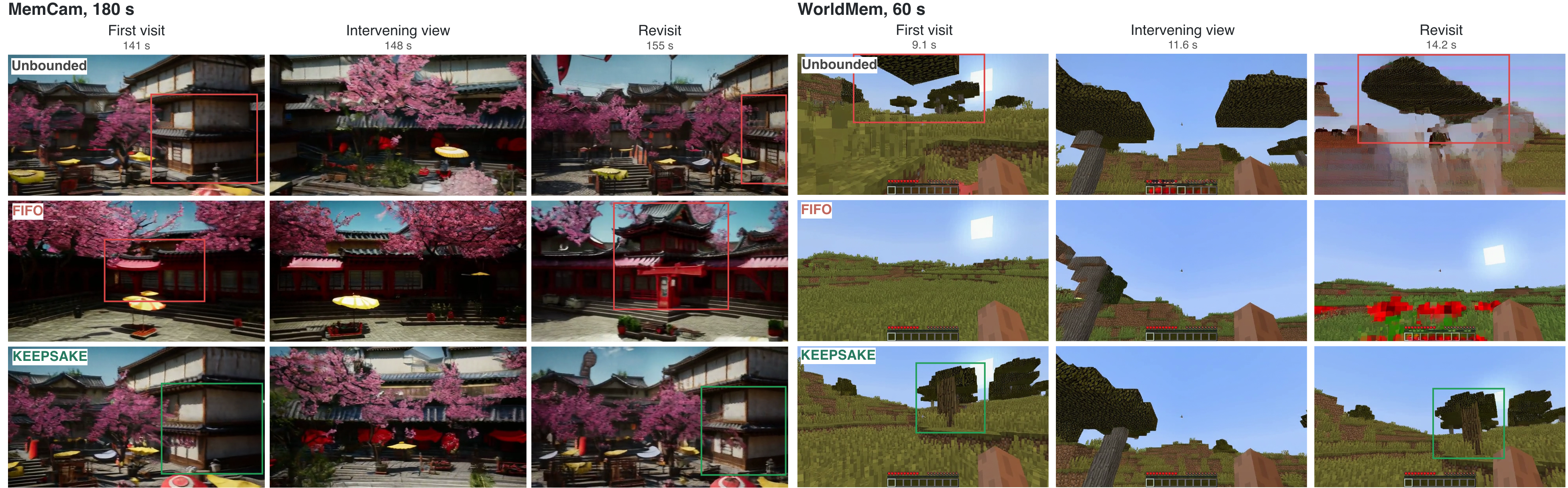}
    \caption{
    \textbf{Qualitative comparison of camera revisits.}
    Examples from 180-second MemCam rollouts (left) and 60-second WorldMem rollouts (right). Rows compare unbounded retention, FIFO, and \Keep{} with $B=32$; columns show the first visit, an intervening view, and the revisit. Highlighted regions track scene content across visits. \Keep{} better preserves the appearance and structure of previously observed regions when the camera returns.}
    \label{fig:archive-diagnostics}
    \vspace{-1.em}
\end{figure}

\paragraph{Longer rollouts.}
We evaluate fifteen matched 180-second MemCam trajectories to
test bounded retention over a longer generation horizon, keeping
the memory budget fixed at $B=32$.
Figure~\ref{fig:long-horizon-extension} reports this evaluation separately
from the 60-second suite. Complete retention stores 5,397
observations by the end of generation, whereas \Keep{} retains
only 32. Despite this reduction, \Keep{} lowers FVD from 734.2
to 476.6 (35.1\%).
The improvement over complete retention therefore extends to
three-minute generation without increasing the memory budget. Visual examples in figure \ref{fig:archive-diagnostics} shows camera revisits maintain consistency with our model.

% Complete replacement for the supplied figure, not an extra nested figure.
% Preamble: \usepackage{pgfplots,subcaption}
%           \pgfplotsset{compat=1.18}
% Self-contained: no external inputs, image files, colors or \Keep macro required.
% Complete replacement for the supplied figure, not an extra nested figure.
% Preamble: \usepackage{pgfplots,subcaption}
%           \pgfplotsset{compat=1.18}
% Self-contained: no external inputs, image files, colors or \Keep macro required.

% table 2
\begin{table}[!t]
\centering
\caption{
\textbf{Round-trip and beyond-context evaluation.} \Keep{} improves most consistency metrics, with larger gains on the more challenging $360^\circ$ round-trip setting.
}
\label{tab:native-benchmarks}

\normalsize
\setlength{\tabcolsep}{5.5pt}
\renewcommand{\arraystretch}{1.08}

\begin{tabular}{@{}l*{6}{r}@{}}
\toprule
& \multicolumn{3}{c}{$90^\circ$, 153 frames}
& \multicolumn{3}{c}{$360^\circ$, 609 frames} \\
\cmidrule(lr){2-4}
\cmidrule(lr){5-7}
Method
& PSNR $\uparrow$ & SSIM $\uparrow$ & LPIPS $\downarrow$
& PSNR $\uparrow$ & SSIM $\uparrow$ & LPIPS $\downarrow$ \\
\midrule
DFoT
& 16.76 & 0.47 & 0.39
& 8.94 & 0.25 & 0.61 \\
GeometryForcing
& 16.57 & 0.49 & 0.35
& 10.07 & 0.40 & 0.57 \\
MemCam
& \textbf{17.83} & 0.51 & 0.357
& 14.81 & 0.42 & 0.50 \\
\midrule
\Keep{}
& 17.66 & \textbf{0.52} & \textbf{0.34}
& \textbf{16.22} & \textbf{0.46} & \textbf{0.43} \\
\bottomrule
\end{tabular}
\end{table}

We restrict the 180-second extension to MemCam. In our setup,
the original WorldMem configuration does not reliably sustain
coherent generation beyond 60 seconds, exhibiting severe visual
degradation. We therefore retain the 60-second prediction horizon
as the stress-test setting for WorldMem and WorldMem + \Keep{}.

\subsection{Budget Sensitivity and Resource Scaling}
\label{sec:exp-budget}

\paragraph{Quality across budgets.}
Table~\ref{tab:ks-budget} reports the available capacity sweeps without
mixing evaluation horizons. In the 180-second MemCam suite, the lowest
reported FVD occurs at $B=16$, while the lowest LPIPS occurs at $B=64$.
WorldMem's reported 60-second LPIPS is lowest at $B=16$. Increasing
capacity therefore does not consistently improve these within-policy
scores. The sweeps characterize a quality--capacity tradeoff; they do
not establish a universal best budget or an intrinsic penalty for
candidate count.

\paragraph{Archive growth and retrieval workload.}
Under complete retention, the candidate pool grows continuously with generation. In MemCam, the mean unbounded candidate count increases from 671.9 in the first 45 seconds to 4,689.5 in the final 45 seconds of a 180-second rollout. In contrast, \Keep{} caps the retained archive at 32 observations after each update while preserving the original retrieval interface and the amount of context supplied to the generator. This substantially reduces retrieval workload as generation proceeds. End-to-end generation time is reported in Table~\ref{tab:end-to-end-generation}. %in appendix.

%\begin{wraptable}{R}{0.5\columnwidth}
\begin{wraptable}[12]{R}{0.5\columnwidth}
    \vspace{-1.5em}
    \centering

    \caption{\textbf{Memory-budget ablation study.}}
    \vspace{-0.8em}
    \label{tab:ks-budget}

    \small
    \setlength{\tabcolsep}{3pt}
    \renewcommand{\arraystretch}{1}

    \resizebox{\linewidth}{!}{%
        \begin{tabular}{@{}crrrr@{}}
            \toprule
            Budget $B$
            & LPIPS $\downarrow$
            & FVD $\downarrow$
            & \shortstack[r]{VBench $\uparrow$}
            & \shortstack[r]{Query time $\downarrow$} \\
            \midrule

            \multicolumn{5}{@{}l}{\textit{MemCam, 180 seconds}} \\
            \addlinespace[2pt]
            16  & 0.5876
                & \textbf{446.1}
                & \textbf{74.90}
                & \textbf{24.80 ms} \\
            32  & 0.5876
                & 476.6 & 74.65 & 43.77 ms \\
            64  & \textbf{0.5865}
                & 493.9 & 73.99 & 105.15 ms \\
            128 & 0.5913
                & 515.5 & 73.82 & 209.78 ms \\

            \midrule
            \multicolumn{5}{@{}l}{\textit{WorldMem, 60 seconds}} \\
            \addlinespace[2pt]
            16  & \textbf{0.525}
                & \textbf{1041.75}
                & \textbf{74.28}
                & \textbf{3.52 ms} \\
            32  & 0.534
                & 1116.92 & 74.05 & 4.54 ms \\
            64  & 0.545
                & 1128.46 & 73.11 & 7.63 ms \\
            128 & 0.577
                & 1601.81 & 71.80 & 12.85 ms \\

            \bottomrule
        \end{tabular}%
    }
\end{wraptable}

\subsection{Retention and Retrieval Diagnostics}
%\subsection{Memory Diagnostics}
\label{sec:exp-mechanism}

\paragraph{Retrieved-memory quality.}
Figure~\ref{fig:retrieval-quality-paired} shows that better viewpoint alignment does not necessarily imply more reliable historical evidence. Selected memories can match the query geometry while exhibiting greater visual drift, indicating that view relevance and memory fidelity are distinct. This distinction is important in long-horizon generation, because a geometrically suitable memory may still propagate degraded scene content. These results motivate evaluating not only which observations are stored, but also the fidelity of the evidence eventually retrieved from them.

\paragraph{Retention versus retrieval failure.}
We further decompose memory failure into two retrospective diagnostics: \emph{retention gap}, measuring useful evidence lost through eviction, and \emph{selection gap}, measuring the retriever's shortfall relative to the best evidence still retained. Figure~\ref{fig:retention-selection} shows full retention removes eviction loss but still exhibits the largest selection gap. In contrast, bounded policies trade a small amount of retention loss for a substantially smaller selection gap. This suggests that a compact archive can make useful evidence easier to retrieve, rather than simply reducing storage.

%\paragraph{View relevance versus historical-image fidelity.}
%The complete-retention diagnostic in Fig.~\ref{fig:retrieval-quality-paired} compares changes in view mismatch with changes in the fidelity of the selected historical content. The reported trend shows better view alignment alongside less faithful stored images. The second panel compares the selected evidence with the best available evidence in the same archive. These observations motivate examining both what memory preserves and what the retriever actually uses; they do not by themselves establish a causal error-propagation chain.

%\paragraph{Retention versus selection.}
%Using ground-truth target features retrospectively, we distinguish \emph{retention gap}, useful evidence lost through eviction, from \emph{selection gap}, the retriever's shortfall relative to the best retained evidence. These are evaluation diagnostics, not online objectives. On fifteen common-source trajectories, Figure~\ref{fig:retention-selection} shows that complete retention has no deletion loss but the largest reported selection gap. This separates two consequences of archive maintenance: preserving useful evidence and making it available to the actual reader are not the same as having that evidence selected.

\section{Discussion and Conclusion}
\label{sec:conclusion}

We introduced \Keep{}, a training-free controller that maintains
fixed-capacity spatial memory by reassessing geometric and visual
substitutability as new views arrive. Across MemCam and WorldMem,
\Keep{} improves FVD and LPIPS over complete retention without
changing the generator or retrieval rule. On fifteen matched
180-second MemCam trajectories, it reduces FVD by 35.1\% while
retaining 32 rather than 5,397 observations. Our diagnostics
further distinguish evidence lost through eviction from retained
evidence left unselected by retrieval, highlighting archive
maintenance as a complement to retrieval. Together, these results
demonstrate that selective retention can improve generation
quality, not merely reduce storage.
\subsection*{AI use statement}

In this work, we have used LLMs to retouch paragraphs. We also used AI tools to revise formatting issues for figures and tables.

\subsection*{Ethics statement}

We do not include any human-related data or anything harmful that might raise ethical issues. Our scope is memory budgeting for long horizon interactive video generation. 

\subsection*{Reproducibility statement}

We include all our memory policy code in the supplementary materials. We also provide full implementation details in the appendix. Our method is reproducible and testable with other papers due to its plug-and-play nature as well.

\bibliography{iclr2027_conference}
\bibliographystyle{iclr2027_conference}

\clearpage
\appendix

\section{Implementation and Evaluation Details}
\label{app:ks-experiments}

\subsection{Evaluation Suites and Matching}
\label{app:ks-suites}

\paragraph{Extended rollouts.}
We evaluate MemCam~\citep{memcam} on
Context-as-Memory~\citep{contextasmemory} and
WorldMem~\citep{worldmem} on its Minecraft world-simulation dataset.
The primary evaluation contains fifteen matched 60-second videos per
system. A separate MemCam evaluation contains fifteen matched
180-second trajectories. Results are reported separately by system
and duration. The MemCam suite also informed policy development.

Within each extended-rollout suite, retention policies share the
initial inputs, prescribed control sequences, generator checkpoint,
generation settings, and random seeds. Only the persistent-memory
update changes. Each policy conditions generation on its own retained
archive, allowing retention decisions to affect subsequent outputs.
The generator and retrieval rule remain unchanged.

\paragraph{Native camera round trips.}
We additionally evaluate MemCam + \Keep{} using MemCam's
camera-round-trip protocol on Context-as-Memory. The two settings
contain $90^\circ$ round trips of 153 frames and $360^\circ$ round
trips of 609 frames. \Keep{} uses $B=32$ in both settings.
This benchmark is reported separately from the extended-rollout suites.
% EDIT: State the evaluated scenes and whether the DFoT,
% GeometryForcing, and MemCam numbers were rerun or taken from the
% original benchmark. The supplied sources do not settle this.

\subsection{Host Configurations and Memory Interfaces}
\label{app:ks-configuration}

Each MemCam and WorldMem generation job was allocated one NVIDIA
H100 PCIe GPU, 16 CPUs, and 64\,GiB of host memory.
The \texttt{memcam} environment uses Python 3.10 and PyTorch 2.5.1
(CUDA 12.4 build), with the \texttt{cuda/cuda-12.1.0} module loaded.
% Local environment: $HOME/.conda/envs/memcam.
% EDIT: Add WorldMem's software versions only after verifying its
% environment; the versions above describe the memcam environment.

MemCam uses Wan2.1 1.3B~\citep{wan} with 76-frame chunks,
50 denoising steps, and $\ell_2$-normalized DINOv2-Base~\citep{dinov2}
retention descriptors. WorldMem uses its released Minecraft
checkpoints at $128\times128$ resolution with 20 DDIM steps
($\eta=0$); retention descriptors are VAE latents pooled to
$4\times4$, flattened, and $\ell_2$-normalized.
\Keep{} preserves native memory representations and retrieval rules:
MemCam retrieves one frame per query for 76 context slots per chunk,
while WorldMem retrieves eight observations per query.

\subsection{Metrics and Reporting}
\label{app:ks-evaluation}

\paragraph{Generation quality.}
LPIPS~\citep{lpips} compares generated frames with ground truth at
corresponding trajectory indices, while FVD~\citep{fvd} measures
video-level distributional quality. Lower values are better for both.
The native round-trip benchmark additionally reports PSNR and SSIM,
for which higher values are better. Within each extended-rollout
suite, policies are evaluated on the same video cohort with the same
metric configuration.

We evaluate standard VBench~\citep{vbench} dimensions: subject
consistency, background consistency, aesthetic quality, and imaging quality. Scores are multiplied
by 100. Together, these dimensions describe perceptual quality and
temporal behavior alongside the reference-based fidelity metrics.
% EDIT: Specify the LPIPS network and preprocessing, the FVD detector
% and clip sampling, and the VBench evaluator version from the scripts.
% The main table reports an Average/Aggregate column, but the draft
% does not define its computation. Insert the verified formula here;
% do not assume it is the mean of the displayed dimensions.

\paragraph{Resource accounting.}
The memory budget counts retained observations, rather than the
context slots filled for an individual generation step. Stored-item
counts describe the archive; they are distinct from GPU memory use,
host-memory allocation, and storage of generated outputs. Retrieval
latency and total generation time are likewise separate quantities.
% EDIT: The timing columns say ms/query, whereas one caption says
% per chunk. Verify the timer's scope and normalize the unit throughout.
% Record timing hardware, CPU threads, warm-up, repetitions, and any
% synchronization from the actual timing code. The generation-job
% resource allocation alone does not establish the lookup-timing setup.

\section{Retention Policies and Update Details}
\label{app:ks-baselines}

\subsection{Retention Baselines}
\label{app:ks-baseline-algorithms}

The primary comparisons use complete retention, FIFO, K-center,
Marginal Coverage Eviction (MCE), and \Keep{}. All bounded policies
use $B=32$ in the main comparison. They operate through the host's
persistent-memory interface without replacing its retrieval rule.

\paragraph{Complete retention.}
All observations admitted by the host remain available for retrieval.
The archive grows as new observations arrive, while the number of
retrieved context slots follows the original host configuration.

\paragraph{FIFO.}
When the archive exceeds the budget, FIFO discards the oldest
observations eligible for eviction until the capacity constraint is
satisfied. Retention is therefore driven by recency rather than
geometric or visual substitutability.

\paragraph{K-center.}
K-center uses geometry and appearance to select representative
observations, targeting the distance between the least-covered
historical view and the retained archive shown in Algorithm \ref{alg:kcenter}. Its objective emphasizes
worst-case coverage rather than recency or the number of substitutes
for an individual observation.

\begin{algorithm}[t]
\caption{Archive-Based K-Center Retention}
\label{alg:kcenter}
\begin{algorithmic}[1]
\Require Candidates $P$, historical archive $H$, budget $B$
\Require Protected frames $F$, distance function $d$
\Ensure Retained bank $S$, with $|S|\leq B$
\If{$|P|\leq B$}
    \State \Return $P$
\EndIf
\If{$H=\emptyset$}
    \State $H\gets P$
\EndIf
\State $S\gets F$
\If{$S=\emptyset$}
    \State $p_0\gets
    \displaystyle\arg\min_{p\in P}
    \frac{1}{|H|}\sum_{h\in H}d(h,p)$
    \State $S\gets\{p_0\}$
\EndIf
\While{$|S|<B$}
    \State $h^\star\gets
    \displaystyle\arg\max_{h\in H}\min_{s\in S}d(h,s)$
    \State $p^\star\gets
    \displaystyle\arg\min_{p\in P\setminus S}d(h^\star,p)$
    \State $S\gets S\cup\{p^\star\}$
\EndWhile
\State \Return $S$
\end{algorithmic}
\end{algorithm}
% EDIT: Insert the implemented distance, initialization, candidate
% and coverage sets, selection schedule, and tie-breaking rule.
% The source describes the objective but not the exact implementation;
% do not substitute an assumed farthest-first algorithm.

\paragraph{Marginal Coverage Eviction.}
MCE uses geometry and appearance to measure collective coverage of
representative historical queries. It repeatedly removes the
observation whose deletion causes the smallest decrease in coverage shown in Algorithm \ref{alg:mce}.
Unlike \Keep{}'s fixed within-update ranking, MCE bases each removal
on the marginal coverage contribution to the retained set.

\begin{algorithm}[]
\caption{Marginal Coverage Eviction (MCE)}
\label{alg:mce}
\begin{algorithmic}[1]
\Require Candidates $P$, budget $B$, protected frames $F$
\Require Queries $Q$, weights $w_q$, similarities $K(q,m)\in[0,1]$
\Ensure Retained bank $S$, with $|S|\leq B$
\State Define coverage:
\[
    C(S)=\sum_{q\in Q}w_q
    \left(1-\prod_{m\in S}(1-K(q,m))\right).
\]
\State $S \gets P$
\While{$|S|>B$}
    \State $m^\star \gets
    \displaystyle\arg\min_{m\in S\setminus F}
    \bigl[C(S)-C(S\setminus\{m\})\bigr]$
    \State $S \gets S\setminus\{m^\star\}$
\EndWhile
\State \Return $S$
\end{algorithmic}
\end{algorithm}
% EDIT: Insert the implemented coverage function, query-set construction
% and capacity, initialization, marginal-score update, and tie-breaking.
% Document auxiliary query storage separately from the retained bank.

\subsection{Memory Updates and Protection}
\label{app:ks-protection}

At update $t$, the retained bank $M_{t-1}$ and incoming observations
$N_t$ form the candidate set $V_t=M_{t-1}\cup N_t$. If
$|V_t|\leq B$, \Keep{} retains all candidates. Otherwise, it removes
$|V_t|-B$ eviction-eligible observations according to the priorities
in Eq.~\ref{eq:retention-utility}.

Protected observations are excluded from eviction but count toward
$B$. They are not additional memory outside the budget. Protection
must therefore leave enough eligible observations to satisfy the
capacity constraint. The resulting archive remains a subset of
$V_t$ in the host's native representation.
% EDIT: Add the current policy-specific protection rules from code.
% An earlier draft lists different initial-frame rules across policies;
% do not claim that all policies use identical protection. Also specify
% the lifetime of endpoint protection for each host.

\subsection{Numerical Conventions and Default Parameters}
\label{app:ks-implementation}

Camera poses are represented as camera-to-world rotations and
translations. Translation distances are normalized by the median
nonzero pairwise translation distance in the current candidate set,
with a floor of $\epsilon=10^{-6}$. When all positions coincide,
the translation scale is set to one. Visual descriptors are
$\ell_2$-normalized, and negative cosine similarities are clipped
to zero when constructing the pose-appearance affinities.

Self-comparisons are excluded from both the substitute count and
maximum similarity. An edge is included when its affinity is at
least $T$, whereas maximum similarity is computed over all other
candidates, including affinities below $T$. Priorities are computed
once over $V_t$ and remain fixed throughout that update's evictions.
Equal-priority ties are broken by earlier insertion. The next update
recomputes priorities using the new candidate set.

Table~\ref{tab:app-parameters} lists the shared parameter settings.
The budget sweep varies $B\in\{16,32,64,128\}$ while holding the
other retention parameters fixed.

\begin{table}[htbp]
    \centering
    \caption{
        \textbf{Default retention settings.}
        The same values are used for MemCam and WorldMem.
    }
    \label{tab:app-parameters}
    \small
    \setlength{\tabcolsep}{6pt}
    \renewcommand{\arraystretch}{1.1}
    \begin{tabular*}{\linewidth}{
        @{\extracolsep{\fill}}llr@{}
    }
        \toprule
        Parameter & Role & Value \\
        \midrule
        $B$ & Retained-observation budget & 32 \\
        $\alpha$ & Pose--appearance weighting & 0.65 \\
        $T$ & Strong-substitute threshold & 0.65 \\
        $\tau$ & Degree saturation threshold & 3 \\
        $\beta$ & Inverse-degree weight & 0.5 \\
        $\lambda$ & Closest-substitute weight & 0.25 \\
        \bottomrule
    \end{tabular*}
\end{table}

\section{Additional Numerical Results and Resource Scaling}
\label{app:ks-additional-results}

\subsection{Long-Horizon Generation}
\label{app:ks-long-horizon}

Table~\ref{tab:app-long-horizon} provides the numerical values
underlying Fig.~\ref{fig:long-horizon-extension}. On the fifteen
matched 180-second MemCam trajectories, \Keep{} improves both
LPIPS and FVD over complete retention and FIFO while maintaining $B=32$.

\begin{table}[htbp]
    \centering
    \caption{
        \textbf{Long-horizon MemCam results.}
        Fifteen matched 180-second trajectories;
        bounded policies use $B=32$.
    }
    \label{tab:app-long-horizon}
    \small
    \setlength{\tabcolsep}{5pt}
    \renewcommand{\arraystretch}{1.1}
    \begin{tabular*}{\linewidth}{
        @{\extracolsep{\fill}}lrrr@{}
    }
        \toprule
        Policy
        & Retained items
        & LPIPS $\downarrow$
        & FVD $\downarrow$ \\
        \midrule
        Complete retention & 5,397 & 0.5980 & 734.2 \\
        FIFO               & 32    & 0.6514 & 677.3 \\
        \Keep{}            & 32    & \textbf{0.5876} & \textbf{476.6} \\
        \bottomrule
    \end{tabular*}
\end{table}

\subsection{Budget Sensitivity}
\label{app:ks-budget-details}

The budget sweep in Table~\ref{tab:ks-budget} evaluates MemCam
at 180 seconds and WorldMem at 60 seconds. In MemCam, the lowest
reported FVD occurs at $B=16$, whereas the lowest LPIPS occurs
at $B=64$. In WorldMem, both metrics are lowest at $B=16$ among
the tested budgets. Retrieval time increases with capacity in both
systems. These results characterize the operating tradeoff of the
retention policy: increasing the budget does not consistently
improve generation quality.
% EDIT: Full standard-VBench dimension tables or component ablations
% can be inserted here when their actual outputs are supplied. The
% current source does not contain those additional numerical results.

\subsection{Archive and Retrieval Scaling}
\label{app:ks-resources}

Let $s$ be the maximum number of observations added per chunk.
Since the retained bank contains at most $B$ observations, a
\Keep{} update processes at most $B+s$ candidates. For fixed
descriptor dimension, direct computation of all pairwise affinities
requires $O((B+s)^2)$ work and temporary storage. Sorting the
priorities adds $O((B+s)\log(B+s))$ work. The persistent archive
requires $O(B)$ storage for fixed observation sizes.

With a fixed number of retrieval queries per chunk and fixed
initial history, complete retention requires $O(ns)$ archive items
and $O(n^2s)$ cumulative exhaustive candidate comparisons over
$n$ chunks. Bounded retention uses at most $B$ archived observations
and $O(nB)$ cumulative comparisons. These bounds describe the
retained archive and candidate-scoring work, not the generator's
weights, activations, or output buffers.

For the 180-second MemCam evaluation, the mean unbounded candidate
count grows from 671.9 in the first 45 seconds to 4,689.5 in the
last 45 seconds. \Keep{} caps the retained bank at 32 after each
update without changing the number of retrieved context slots.

\begin{table}[t]
\centering
\caption{
End-to-end generation time per completed video.
Comparisons are made within each system.
}
\label{tab:end-to-end-generation}
\small
\begin{tabular}{llrrrr}
\toprule
System & Rollout
& Unbounded
& \Keep{} B32
& Reduction
& Speedup \\
\midrule
MemCam   & 60 s  & 106.46 min & \textbf{65.46 min}
& \textbf{41.00 min (38.5\%)} & \textbf{1.626$\times$} \\
MemCam   & 180 s & 587.99 min & \textbf{193.14 min}
& \textbf{394.85 min (67.2\%)} & \textbf{3.044$\times$} \\
WorldMem & 60 s  & 12.30 min & \textbf{11.85 min}
& \textbf{0.45 min (3.7\%)} & \textbf{1.038$\times$} \\
\bottomrule
\end{tabular}
\end{table}
\section{Retention and Retrieval Diagnostics}
\label{app:ks-diagnostics}

\subsection{Generation Evaluations and Common-Source Diagnostics}
\label{app:ks-diagnostic-protocol}

In the generation evaluations, each policy produces its own video
while updating and querying its own archive. The resulting scores
therefore reflect the interaction between retention and subsequent
generation. The common-source diagnostics instead load
policy-selected historical images from the same complete-retention
rollout. This separates differences in the selected historical
evidence from differences between independently generated videos.

Ground-truth references are used retrospectively to evaluate the
selected evidence. They are not inputs to the online retention
priority, which uses camera poses and appearance descriptors from
the observations currently available to the controller.

\subsection{View Relevance and Historical-Image Fidelity}
\label{app:ks-memory-fidelity}

Figure~\ref{fig:retrieval-quality-paired} distinguishes view mismatch
from memory corruption. The former concerns agreement with the
queried viewpoint, whereas the latter concerns the fidelity of the
selected historical content. Both are measured using DINOv2 cosine
distance in the fifteen-trajectory, 60-second common-source
comparison. Table~\ref{tab:app-memory-fidelity} gives the means and
95\% trajectory-bootstrap confidence intervals shown in the figure.
% EDIT: Insert the exact image/feature pairs defining each distance
% from the diagnostic script. The manuscript does not specify them
% precisely enough to reconstruct the equations. Also document the
% aggregation and bootstrap implementation rather than assuming them.

\begin{table}[htbp]
    \centering
    \caption{
        \textbf{Common-source memory diagnostics.}
        Fifteen 60-second trajectories; bounded policies use $B=32$.
        Entries are means with 95\% trajectory-bootstrap confidence
        intervals. Lower distances are better.
    }
    \label{tab:app-memory-fidelity}
    \small
    \setlength{\tabcolsep}{5pt}
    \renewcommand{\arraystretch}{1.15}
    \begin{tabular*}{\linewidth}{
        @{\extracolsep{\fill}}lrr@{}
    }
        \toprule
        Policy & View mismatch & Memory corruption \\
        \midrule
        Complete retention
        & 0.1256 [0.1041, 0.1479]
        & 0.4871 [0.4012, 0.5716] \\
        FIFO
        & 0.1557 [0.1336, 0.1777]
        & 0.5022 [0.4195, 0.5835] \\
        \Keep{}
        & 0.3254 [0.2579, 0.3964]
        & 0.4079 [0.3159, 0.4989] \\
        \bottomrule
    \end{tabular*}
\end{table}

\Keep{} has higher mean view mismatch but lower mean memory
corruption than complete retention in this comparison. The two
quantities therefore need to be distinguished: closer agreement
with a query viewpoint does not necessarily coincide with higher
historical-image fidelity.

\subsection{Retention Gap and Selection Gap}
\label{app:ks-gap-definitions}

The retention-selection diagnostic separates two sources of
shortfall relative to useful historical evidence. Retention gap
measures evidence lost through eviction by comparing the best
available full-history evidence with the best evidence still
retained. Selection gap measures the retriever's shortfall relative
to the best retained evidence. The comparison uses target
ground-truth features retrospectively; neither gap is an online
retention objective.
% EDIT: Add the exact gap equations, candidate eligibility rules,
% feature definition, and query/trajectory aggregation from the code.
% The plotting-source comment identifies 13 trajectories at 180 s,
% whereas the current caption says 15. Resolve this before assigning
% a cohort count to this subsection or table.

Table~\ref{tab:app-gap-values} records the rounded coordinates in
Fig.~\ref{fig:retention-selection}. Complete retention has zero
retention gap by construction and a selection gap of 0.2188.
At each displayed budget, \Keep{} has a smaller retention gap
than FIFO, while FIFO has a smaller selection gap. Increasing
capacity reduces retention gap but increases selection gap along
both plotted policy curves. The diagnostic thus separates
preserving evidence from having the existing retriever select it.

\begin{table}[htbp]
    \centering
    \caption{
        \textbf{Retention-selection diagnostic at 180 seconds.}
        Rounded coordinates from Fig.~\ref{fig:retention-selection}.
        Complete retention has retention gap 0 and selection gap 0.2188.
    }
    \label{tab:app-gap-values}
    \small
    \setlength{\tabcolsep}{5pt}
    \renewcommand{\arraystretch}{1.1}
    \begin{tabular*}{\linewidth}{
        @{\extracolsep{\fill}}crrrr@{}
    }
        \toprule
        & \multicolumn{2}{c}{FIFO}
        & \multicolumn{2}{c}{\Keep{}} \\
        \cmidrule(lr){2-3}
        \cmidrule(lr){4-5}
        Budget $B$
        & Retention gap & Selection gap
        & Retention gap & Selection gap \\
        \midrule
        16  & 0.1872 & 0.0448 & 0.0669 & 0.1037 \\
        32  & 0.1668 & 0.0646 & 0.0426 & 0.1470 \\
        64  & 0.1429 & 0.0867 & 0.0366 & 0.1596 \\
        128 & 0.1203 & 0.1084 & 0.0298 & 0.1783 \\
        \bottomrule
    \end{tabular*}
\end{table}

\subsection{Sensitivity of the Historical-Fidelity Comparison}
\label{app:ks-common-source}

For the PSNR and SSIM analysis, selected historical images are
loaded from the same complete-retention rollout and compared with
ground truth at their own historical indices. This evaluates the
fidelity of stored content rather than agreement with the current
query viewpoint. Across fifteen trajectories, the aggregate
\Keep{}-minus-complete-retention differences are $+4.629$\,dB
PSNR and $+0.1512$ SSIM.

Table~\ref{tab:app-fidelity-sensitivity} shows how these differences
change under two successive restrictions. First, we exclude
selection disagreements in which \Keep{} chooses the initial
conditioning image. We then additionally require both selected
observations to have field-of-view overlap of at least 0.80.

\begin{table}[htbp]
    \centering
    \caption{
        \textbf{Sensitivity of historical-image fidelity.}
        Differences are \Keep{} minus complete retention on common-source
        images, evaluated against ground truth at their historical indices.
        The restrictions are cumulative.
    }
    \label{tab:app-fidelity-sensitivity}
    \small
    \setlength{\tabcolsep}{5pt}
    \renewcommand{\arraystretch}{1.15}
    \begin{tabularx}{\linewidth}{
        @{}>{\raggedright\arraybackslash}Xrr@{}
    }
        \toprule
        Comparison subset
        & $\Delta$PSNR (dB)
        & $\Delta$SSIM \\
        \midrule
        Aggregate comparison
        & +4.629 & +0.1512 \\
        Exclude initial-image selection disagreements
        & +1.060 & +0.0449 \\
        Additionally require FOV overlap $\geq0.80$ for both selections
        & +0.103 & +0.0098 \\
        \bottomrule
    \end{tabularx}
\end{table}
% EDIT: Add subset sizes and the actual confidence-interval endpoints
% from the diagnostic output. Do not infer them from the means.

Both confidence intervals for the final restricted comparison
include zero. The aggregate advantage therefore depends on the
selection distribution and does not establish a general ability to
identify higher-fidelity memories at matched viewpoints.

\section{Qualitative Comparisons}
\label{app:ks-qualitative}

\begin{figure}[!htbp]
    \centering

    % ---------- (a) Top ----------
    \begin{subfigure}{\linewidth}
        \centering
        \includegraphics[
            width=\linewidth,
            height=0.25\textheight,
            keepaspectratio
        ]{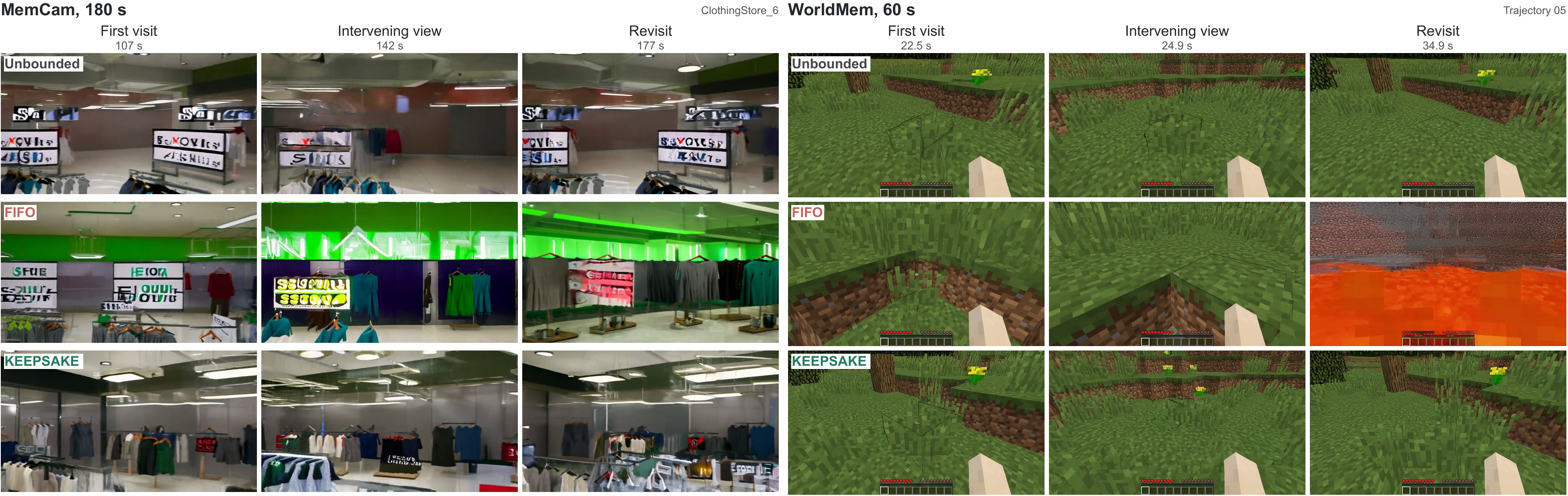}
        \caption{}
        \label{fig:retrieval_fail_a}
    \end{subfigure}

    \par\medskip

    % ---------- (b) Middle ----------
    \begin{subfigure}{\linewidth}
        \centering
        \includegraphics[
            width=\linewidth,
            height=0.25\textheight,
            keepaspectratio
        ]{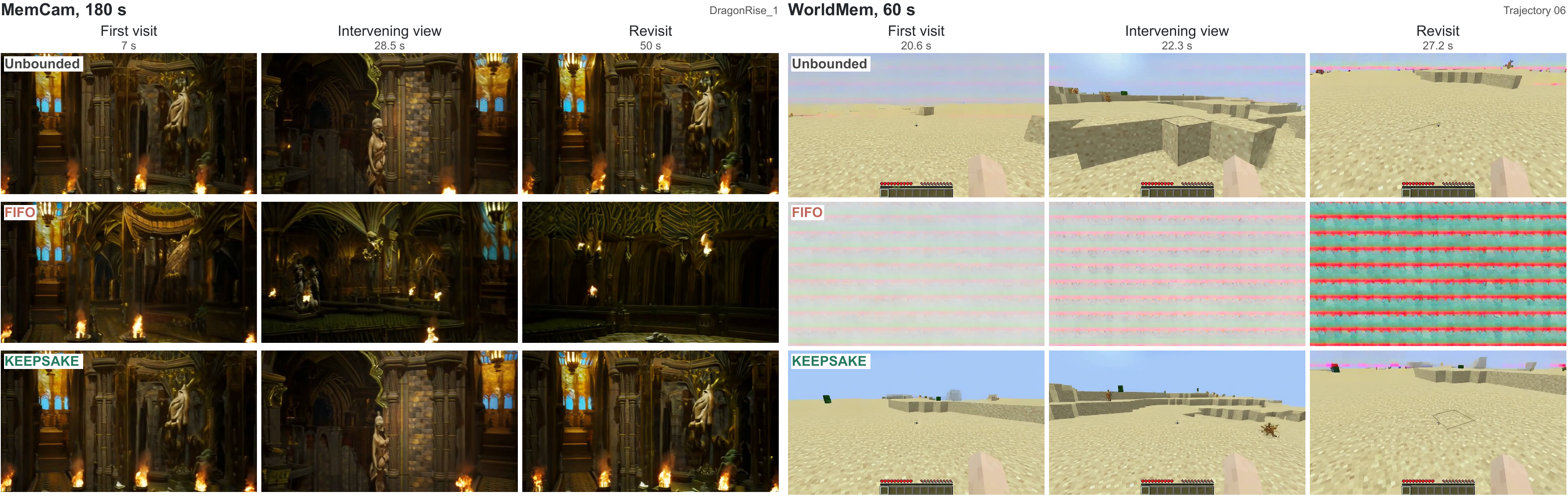}
        \caption{}
        \label{fig:retrieval_fail_b}
    \end{subfigure}

    \par\medskip

    % ---------- (c) Bottom ----------
    \begin{subfigure}{\linewidth}
        \centering
        \includegraphics[
            width=\linewidth,
            height=0.25\textheight,
            keepaspectratio
        ]{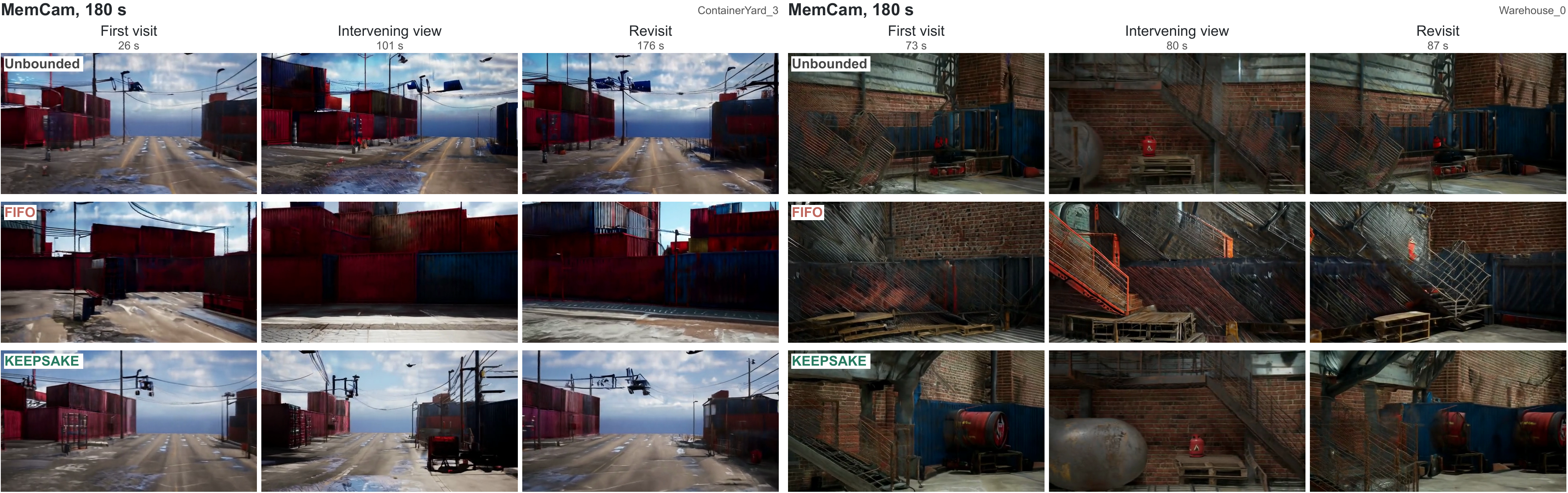}
        \caption{}
        \label{fig:retrieval_fail_c}
    \end{subfigure}

    \caption{\textbf{Additional camera-revisit examples.}}
    \label{fig:rollout}
\end{figure}

\begin{figure}[h]
    \centering
    \includegraphics[width=\linewidth]{
        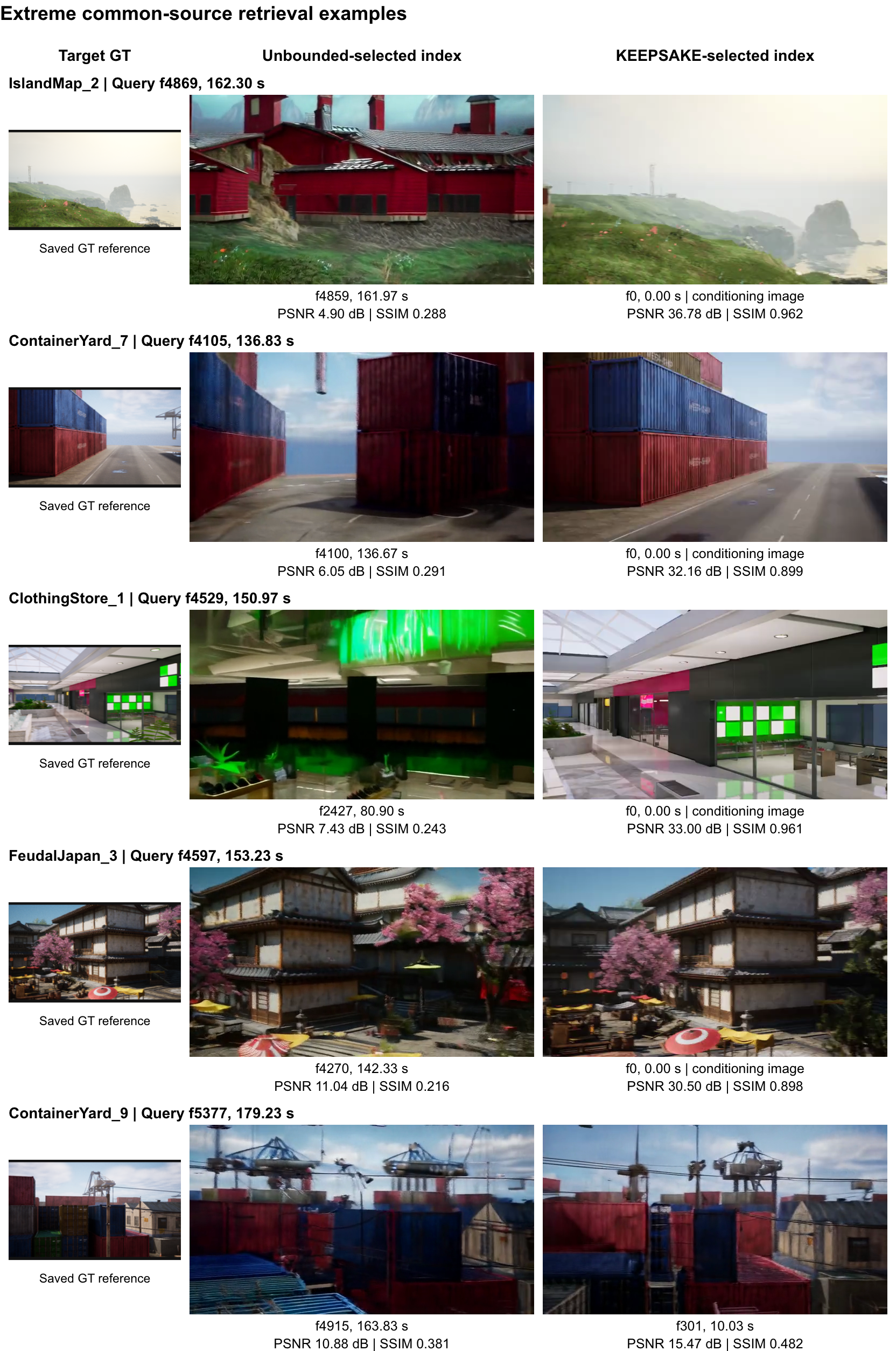
    }
    \caption{
        Selected common-source retrieval examples. For identical queries, unbounded retention and \Keep{} can retrieve substantially different historical observations.
    }
    \label{fig:retrieval_fail}
\end{figure}

We provide two sets of qualitative comparisons. Figure \ref{fig:rollout} presents additional camera-revisit examples comparing unbounded retention, FIFO, and \Keep{}. Figure \ref{fig:retrieval_fail} presents selected common-source queries for which unbounded retention and \Keep{} retrieve different historical observations.

% Optional additional grids. These blocks remain commented until
% the corresponding images have been prepared. Use matched trajectories,
% timestamps, and crop regions; edit captions to match the actual grids.
%
% Requires \usepackage{graphicx} and \usepackage{subcaption}

% \begin{figure}[t]
%     \centering
%     \includegraphics[width=\linewidth]{
%         figures/appendix_worldmem_revisits.pdf
%     }
%     \caption{
%         \textbf{Additional WorldMem revisit comparisons.}
%         Examples from 60-second rollouts. Rows compare complete
%         retention, FIFO, and \Keep{}; bounded policies use $B=32$.
%         Columns show the first visit, an intervening view, and the
%         revisit at the indicated timestamps. Boxes identify
%         corresponding scene regions.
%     }
%     \label{fig:app-worldmem-revisits}
% \end{figure}

\end{document}

%% file: math_commands.tex
\usepackage{amsmath,amsfonts,bm}

\def\eqref#1{equation~\ref{#1}}
\def\1{\bm{1}}

\DeclareMathAlphabet{\mathsfit}{\encodingdefault}{\sfdefault}{m}{sl}
\SetMathAlphabet{\mathsfit}{bold}{\encodingdefault}{\sfdefault}{bx}{n}